\documentclass[runningheads]{llncs}
\usepackage{graphicx}
\usepackage{tikz}
\usepackage{comment}
\usepackage{amsmath,amssymb} % define this before the line numbering.
\usepackage{color}

\usepackage[accsupp]{axessibility}  % Improves PDF readability for those with disabilities.

\usepackage{subfigure}
\usepackage{booktabs}
\usepackage[colorlinks,
            linkcolor=blue,
            anchorcolor=blue,
            citecolor=blue]{hyperref}

\usepackage{orcidlink}
\begin{document}
\pagestyle{headings}
\mainmatter
\def\ECCVSubNumber{123}  % Insert your submission number here

\title{Privacy-Preserving Face Recognition with Learnable Privacy Budgets in Frequency Domain} % Replace with your title

\begin{comment}
\titlerunning{ECCV-22 submission ID \ECCVSubNumber} 
\authorrunning{ECCV-22 submission ID \ECCVSubNumber} 
\author{Anonymous ECCV submission}
\institute{Paper ID \ECCVSubNumber}
\end{comment}
%******************

% \begin{comment}
% \titlerunning{DCTDP}
\titlerunning{Privacy-Preserving Face Recognition with Learnable Privacy Budgets}
\author{Jiazhen Ji\inst{1}\orcidlink{0000-0003-2708-9319} \and Huan Wang\inst{2}\orcidlink{0000-0001-9384-4103} \and Yuge Huang\inst{1}\orcidlink{0000-0001-5387-5992} \and Jiaxiang Wu\inst{1}\orcidlink{0000-0001-9203-9372} \and Xingkun Xu\inst{1}\orcidlink{0000-0001-6399-3415} \and Shouhong Ding\inst{1}\orcidlink{0000-0002-3175-3553} \and ShengChuan Zhang\inst{2}\orcidlink{0000-0002-0800-0609} \and Liujuan Cao\inst{2} \and  Rongrong Ji\inst{2}}
\authorrunning{J. Ji, H. Wang, Y. Huang, and et al.}
\institute{Youtu Lab, Tencent \and
Xiamen University\\
\email{\{royji, yugehuang, willjxwu, xingkunxu, ericding\}@tencent.com, hanawh@stu.xmu.edu.cn, \{zsc\_2016, caoliujuan, rrji\}@xmu.edu.cn}}
%
%******************
\maketitle

\begin{abstract}
Face recognition technology has been used in many fields due to its high recognition accuracy, including the face unlocking of mobile devices, community access control systems, and city surveillance. 
As the current high accuracy is guaranteed by very deep network structures, facial images often need to be transmitted to third-party servers with high computational power for inference.
However, facial images visually reveal the user's identity information. 
In this process, both untrusted service providers and malicious users can significantly increase the risk of a personal privacy breach.  
Current privacy-preserving approaches to face recognition are often accompanied by many side effects, such as a significant increase in inference time or a noticeable decrease in recognition accuracy.
This paper proposes a privacy-preserving face recognition method using differential privacy in the frequency domain. 
Due to the utilization of differential privacy, it offers a guarantee of privacy in theory.
Meanwhile, the loss of accuracy is very slight.
This method first converts the original image to the frequency domain and removes the direct component termed DC.
Then a privacy budget allocation method can be learned based on the loss of the back-end face recognition network within the differential privacy framework.
Finally, it adds the corresponding noise to the frequency domain features.
Our method performs very well with several classical face recognition test sets according to the extensive experiments. Code will be available at \url{https://github.com/Tencent/TFace/tree/master/recognition/tasks/dctdp}.
\end{abstract}

\section{Introduction}

\begin{figure}[t]
\centering\includegraphics[width=0.73\columnwidth]{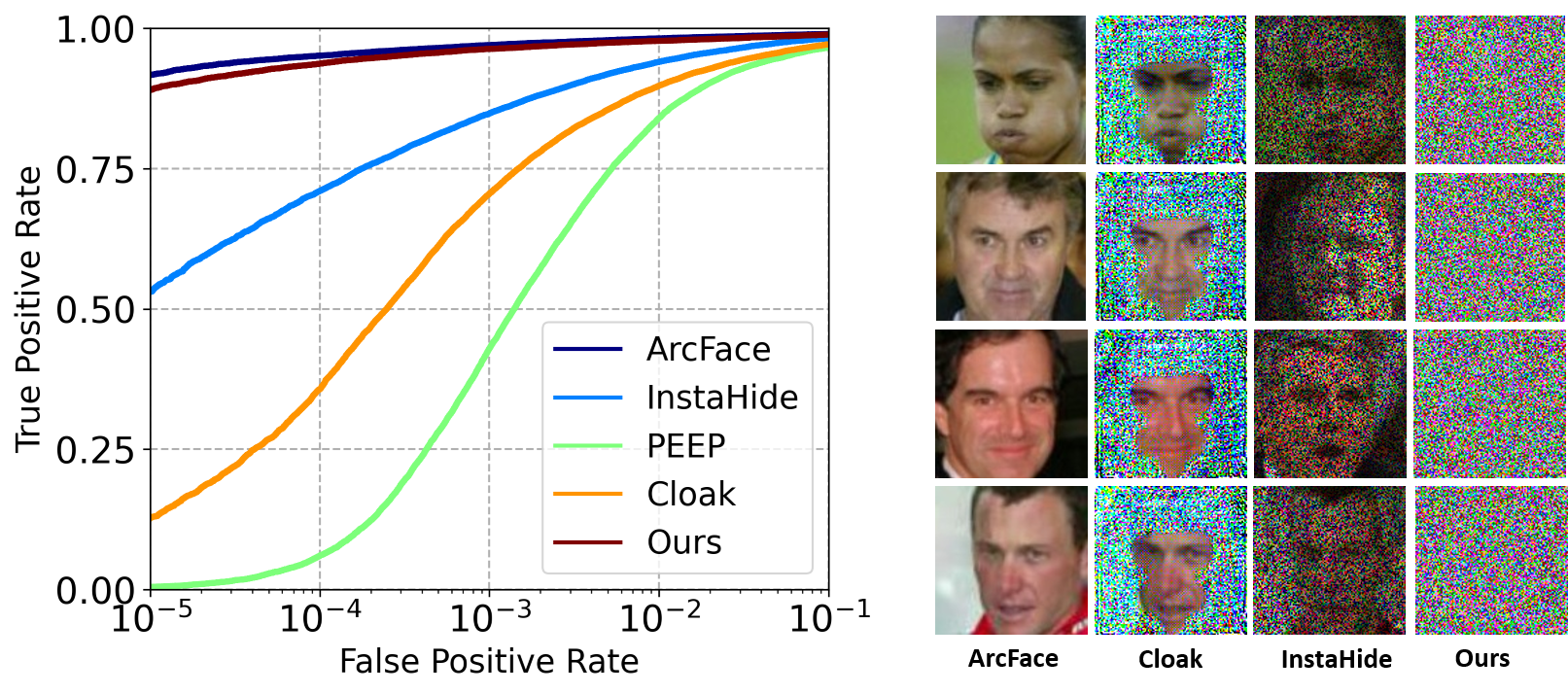}
\caption{\small \textbf{Comparison with different privacy-preserving methods.}
\textbf{Left}: ROC curves under IJB-C dataset, the closer to the upper left area, the better the performance.
\textbf{Right}: Visualization of processed images using LFW dataset as examples. Compared with other methods, our method has good privacy-preserving performance.
}
\label{figfirst}
\end{figure}

With the rapid development of deep learning, face recognition models based on convolutional neural networks have gained a remarkable breakthrough in recent years.
The extremely high accuracy rate has led to its application in many daily life scenarios. 
However, due to the privacy sensitivity of facial images and the unauthorized collection and use of data by some service providers, people have become increasingly concerned about the leakage of their face privacy. 
In addition to these unregulated service providers, malicious users and hijackers pose a significant danger to privacy leakage. 
It is necessary to apply some privacy-preserving mechanisms to face recognition.

Homomorphic encryption \cite{gentry2011implementing} is an encryption method that encrypts the original data and performs inference on the encrypted data. It protects data privacy and maintains a high level of recognition accuracy.
However, the introduction of the encryption process requires a great additional computation.
Therefore, it is not suitable for large-scale and interactive scenarios.

PEEP \cite{chamikara2020privacy} is a very typical approach to privacy protection that makes use of differential privacy. It first converts the original image to a projection on the eigenfaces. Then, it adds noise to it by utilizing the concept of differential privacy to provide better privacy guarantees than other privacy models. 
PEEP has a low computational complexity, not to slow down the inference speed. 
However, it significantly reduces the accuracy of face recognition.
As shown in Fig.\ref{figfirst}, the current face recognition privacy-preserving methods all perform poorly with large data sets and have relatively mediocre privacy-preserving capabilities.

This paper aims to limit the face recognition service provider to learn only the classification result (e.g., identity) with a certain level of confidence but does not have access to the original image (even by some automatic recovery techniques).
We propose a privacy-preserving framework to tackle the privacy issues in deep face recognition.
Inspired by frequency analysis performing well in selecting the signal of interest, we deeply explored the utility of frequency domain privacy preservation.
We use block discrete cosine transform (DCT) to transfer the raw facial image to the frequency domain.
It is a prerequisite for separating information critical to visualization from information critical to identification.
Next, we remove the direct component (DC) channel because it aggregates most of the energy and visualization information in the image but is not essential for identification.
Meanwhile, we consider that elements at different frequencies of the input image do not have the same importance for the identification task. 
Therefore it is not reasonable to set precisely the same privacy budget for all elements. 
We propose a method taking into account the importance of elements for identification.
It only needs to set the average privacy budget to obtain the trade-off between privacy and accuracy. 
Then the distribution of privacy budgets over all the elements will be learned according to the loss of the face recognition model.
Compared with PEEP, our approach to switching to the frequency domain is simpler, faster, and easier to deploy. 
Moreover, because we learn different privacy budgets for different features from the loss of face recognition, our recognition accuracy is also far better than SOTAs.
The contributions of this paper are summarized as follows:
\begin{itemize}
\item We propose a framework for face privacy protection based on the differential privacy method. The method is fast and efficient and adjusts the privacy-preserving capability according to the choice of privacy budget.
\item We design a learnable privacy budget allocation structure for image representation in the differential privacy framework, which can protect privacy while reducing accuracy loss.
\item We design various privacy experiments that demonstrate the high privacy-preserving capability of our approach with marginal loss of accuracy.
\item Our method can transform the original face recognition dataset into a privacy-preserving dataset while maintaining high availability.
\end{itemize}

\section{Related Work}
\subsection{Face Recognition}
The state-of-the-art (SOTA) research on face recognition mainly improve from the perspective of softmax-based loss function which aims to maximize the inter-class discrepancy and minimize the intra-class variance for better recognition accuracy \cite{deng2019arcface,liu2017sphereface,schroff2015facenet,wang2020mis,huang2020curricularface}.
However, existing margin-based methods do not consider potential privacy leakage. 
In actual applications, raw face images of users need to be delivered to remote servers with GPU devices, which raises the user’s concern about the abuse of their face images and potential privacy leakage during the transmission process.
Our method takes masked face images rather than raw face images as the face recognition model’s inputs, which reduces the risk of misuse of user images.
\subsection{Frequency Domain Learning}
Frequency analysis has always been widely used in signal processing, which is a powerful tool for filtering signals of interest.
In recent years, Some works that introduced frequency-domain analysis have been proposed to tackle the various aspects of the problem.

% ~\cite{ehrlich2019deep} proposed a general network learning in the JPEG transform domain that allows the network to squeeze more performance.
For instance,~\cite{gueguen2018faster} trained CNNs directly on the blockwise discrete cosine transform (DCT) coefficients.~\cite{xu2020learning} made an in-depth analysis of the selection of DCT coefficients on three different high-level vision tasks such as image classification, detection, and segmentation.~\cite{huang2021fsdr} presented a frequency space domain randomization technique that achieved superior segmentation performance.
In the field of deepfake, frequency is an essential tool to distinguish real from synthetic images (or videos) \cite{frank2020leveraging}.
As for face recognition,~\cite{kagawade2021fusion} presents a new approach through a feature-level fusion of face and iris traits extracted by polar fast fourier transform.~\cite{wang2022privacy} splits the frequency domain channels of the image and selects only some of them for subsequent tasks.
On the basis of these works, we firstly and deeply explored the utility of privacy-preserving in the frequency domain.
\subsection{Privacy Preserving}
Privacy-preserving research can be broadly categorized based on how to process input data, i.e., encryption or perturbation.
The majority of data encryption methods fall under the Homomorphic Encryption (HE) and Secure Multiparty Computation (SMC)~\cite{boemer2020mp2ml,gilad2016cryptonets,juvekar2018gazelle,liu2017oblivious}.
However, these data encryption methods are unsuitable for current SOTA face recognition systems due to their prohibitive computation cost.
Data perturbation, in contrast, avoids the high cost of encryption by applying a perturbation to raw inputs.
Differential privacy (DP) is a common-used perturbation approach.
\cite{chamikara2020privacy} applied perturbation to eigenfaces utilizing differential privacy that equally split the privacy budget to every eigenface resulting in a great loss of accuracy.
In addition to DP, there are other methods for data disturbance.
For instance,~\cite{huang2020instahide} used the Mixup~\cite{zhang2018mixup} method to perturb data while it has been hacked successfully~\cite{carlini2020attack}.
~\cite{mireshghallah2021not} presented a Gaussian noise disturbance method to suppress unimportant pixels before sending them to the cloud.
Recently, some privacy-preserving methods have also appeared in the field of face recognition.
\cite{bai2021federated} proposed federated face recognition to train face recognition models using multi-party data via federated learning to avoid privacy risks.
K-same~\cite{newton2005preserving} is a de-identification approach by using K-anonymity for face images.
Our method maintains accuracy to the greatest extent based on a learnable privacy budget.

\section{Method}

\begin{figure*}[t!]
\begin{center}
\includegraphics[width=0.85\textwidth]{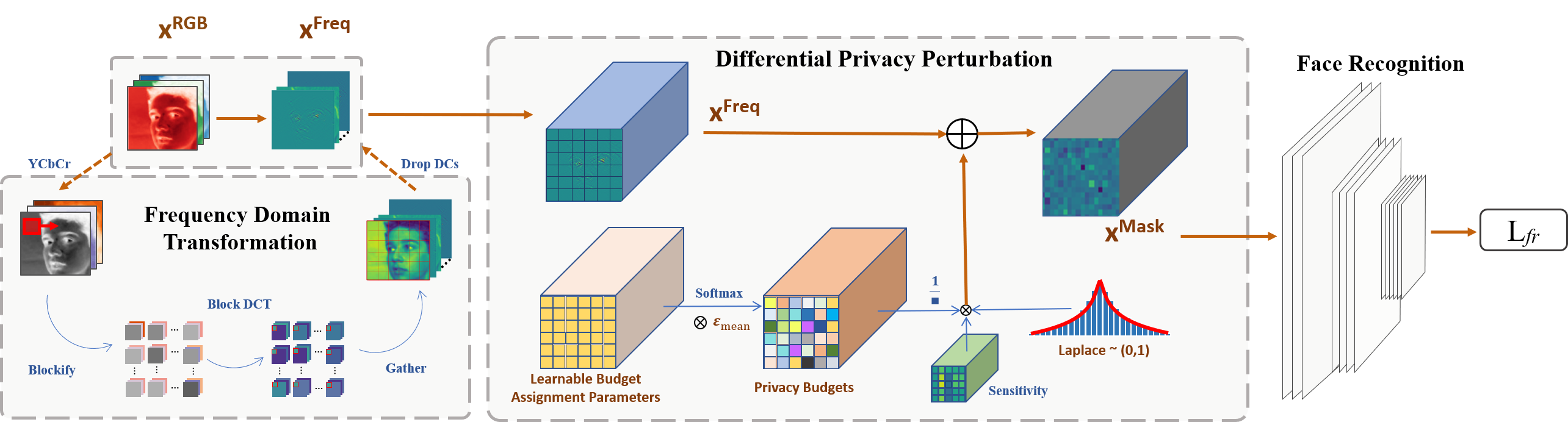}
\end{center}
   \caption{Overview of our proposed method. It consists of three modules: frequency-domain transformation, differential privacy perturbation, and face recognition. The frequency-domain transformation module transforms the facial image to frequency domain features. The differential privacy perturbation module adds perturbations to frequency domain features. The face recognition module takes the perturbed features as input and performs face recognition.
   }
\label{figmodel}
\end{figure*}

\noindent In this section, we describe the framework of our proposed privacy-preserving face recognition method. 
Our method consists of three main modules, a frequency domain transformation module, a perturbation module that utilizes differential privacy, and a face recognition module, as shown in Fig.\ref{figmodel}. 
Each input image is first converted to frequency domain features by the frequency domain transformation module. 
Furthermore, the differential privacy perturbation module will generate the corresponding noise and add it to the frequency domain features. 
Finally, the perturbed frequency domain features will be transferred to the face recognition model. 
Because the size of the perturbed frequency domain features is $[H, W, C]$, we only need to change the input channels of the face recognition model from 3 to $C$ to suit our input. 
In Section \ref{Frequency}, we describe in detail the specific process of frequency-domain conversion. 
In Section \ref{DPback}, we first introduce the background knowledge of differential privacy and then describe the specific improvements in our differential privacy perturbation module.

\subsection{Frequency-Domain Transformation Module}
\label{Frequency}
\noindent
\cite{wang2020high} discovered that humans rely only on low-frequency information for image recognition while neural networks use low-and high-frequency information. 
Therefore, using low-frequency information as little as possible is very effective for image privacy protection.
DCT transformation can be beneficial in separating the low-frequency information that is important for visualization from the high-frequency information that is important for identification. 
In the frequency domain transformation module,  inspired by the compression operation in JPEG, we utilized block discrete cosine transform (BDCT) as our basis of frequency-domain transformation. 
For each input image, we first convert it from RGB color spaces to YCbCr color spaces. 
We then adjust its value range to $[-128, 127]$ to meet the requirement of BDCT input and then split it into $\frac{H}{8} \times \frac{W}{8}$ blocks with a size of 8 $\times$ 8. 
For a fairer comparison and as little adjustment as possible to the structure of the recognition network, we perform an 8-fold up-sampling on the facial images before BDCT. 
Then a normalized, two-dimensional type-II DCT is used to convert each block into 8 $\times$ 8 frequency-domain coefficients. 

\begin{figure}[t!]
\centering
\includegraphics[width=0.75\columnwidth]{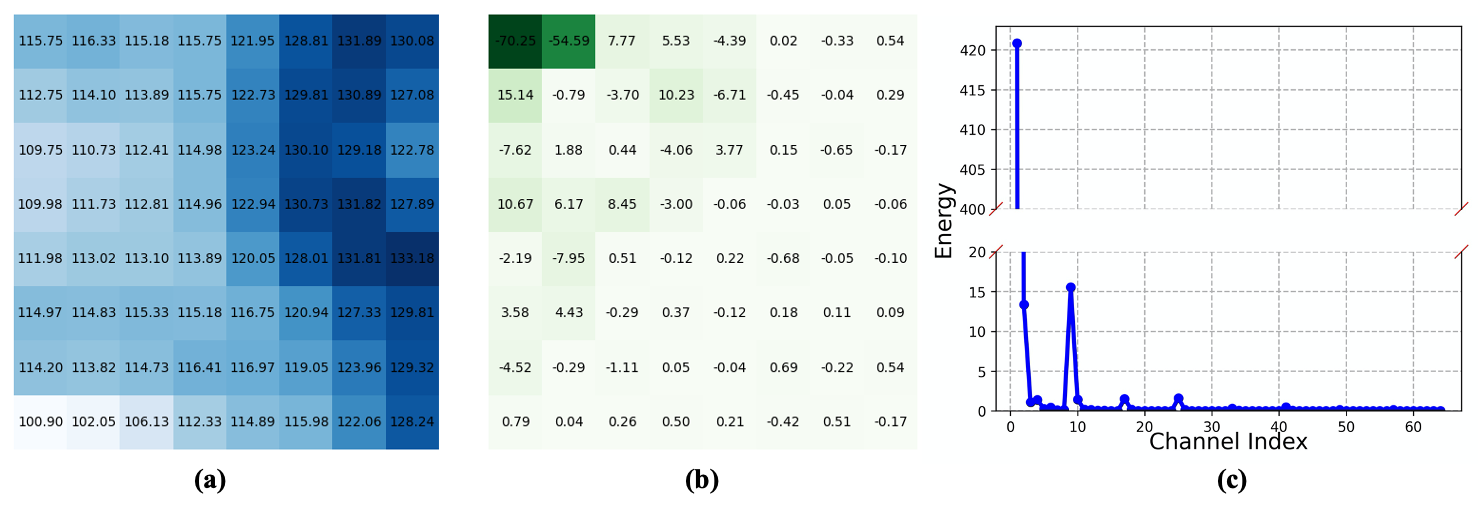} 
\caption{(a) One block of the Original image with size $8\times8$. (b) Frequency-domain features of the image block after DCT transformation. (c) Energies among different channels. DC channels take up most of the energy}
\label{figdct}
\end{figure}

At this time, we can see that the element in the upper left corner in Fig.\ref{figdct}(b) has an extreme value. It defines the basic tone of the entire block, which we call the DC component. 
For elements in the same position in each block, we collect them together in the same channel and arrange them according to the relative position of the block in the original image. 
Here we have converted the original image with size $[H, W, 3]$ into a frequency domain representation with size $[H, W, 8, 8, 3]$. As we can see in Fig.\ref{figdct}(c), the most energy (91.6\%) of facial images is concentrated in the DC channel. 
According to our experiment results in Tab.\ref{tableaccabl}, DC is crucial for visualization, while it has little impact on recognition. 
Thus it is removed before it enters the next module.

\subsection{Differential Privacy (DP)}
\label{DPback}
\subsubsection{Definition.} 
DP \cite{mcsherry2007mechanism} is known to be a privacy model that provides maximum privacy by minimizing the possibility of personal records being identified.
By adding noise to the query information of the database, DP helps to ensure the security of personal information while obtaining comprehensive statistical information. 
For any two databases that differ by only one individual, we say they are adjacent.

A randomized algorithm $\mathcal{A}$ satisfies $\epsilon$-DP if for every adjacent pairs of database $D_1$ and $D_2$ and all $\mathcal{Q} \subseteq  Range(\mathcal{A})$, Equation \ref{eqldp} holds.
\begin{equation}
Pr(\mathcal{A}(D_1) \in \mathcal{Q}) \leq e^\epsilon Pr(\mathcal{A}(D_2) \in \mathcal{Q})
\label{eqldp}    
\end{equation}

\subsubsection{DP for Generative Models.}
Since our goal is to protect facial privacy by scrambling the images after BDCT, we need to move from the database domain to the generative model representations domain. 
Under the generative model representation domain, query sensitivity and adjacency in the database domain are no longer applicable. 
Each input image should correspond to a separate database. 
Some methods \cite{croft2021obfuscation} apply DP generalization to arbitrary secrets, where a secret is any numerical representation of the data. 

In our approach, we consider the BDCT representation of the facial image as a secret. 
The distance between secrets replaces the notion of adjacency between databases. 
We can control the noise by the distance metric to make similar (in visualization) secrets indistinguishable while keeping very different secrets that remain distinguishable.
Thus, the recoverability is minimized while ensuring as much identifiability as possible.
Therefore the choice of distance metric for secrets is critical. 
In our approach, each image is transformed to a BDCT representation with size $[H, W, C]$. Let $R_{i,j,k} = [r^{i,j,k}_{min}, r^{i,j,k}_{max}]$ be the sensitivity of the element in the $[i, j, k]$ position of representation. 
Then we define an element-wise distance as follows:

\begin{equation}
d_{i,j,k}(x_1, x_2) = \frac{|x_1 - x_2|}{r^{i,j,k}_{max} - r^{i,j,k}_{min}} ~~~~~~\forall x_1, x_2  \in R_{i,j,k}
\label{bitdis}    
\end{equation}

Moreover, the distance between the whole representations are defined as follow:

\begin{equation}
\begin{aligned}
d(X_1, X_2) &= \max_{i,j,k}(d_{i,j,k}(x_1, x_2)) \\
\forall X_1, &X_2  \in \mathbb{R}^{H,W,C}
\end{aligned}
\label{wholedis} 
\end{equation}

Thus, the $\epsilon$-DP protection can be guaranteed for mechanism $\mathcal{K}$ if for any representation $X_1$, $X_2$ and $E \in  \mathbb{R}^{H,W,C}$. Equation \ref{newdp} satisfied:

\begin{equation}
\begin{aligned}
Pr(\mathcal{K}(X_1) = E) &\leq e^{\epsilon d(X_1,X_2)} Pr(\mathcal{K}(X_2) = E)\\
\forall X_1, &X_2, E \in \mathbb{R}^{H,W,C}
\end{aligned}
\label{newdp} 
\end{equation}
With the definition of distance between representations and DP for representations, we can claim

\begin{lemma}
\label{lemma}
Any BDCT representation of image $X \in \mathbb{R}^{H,W,C}$ can be protected by $\epsilon$-DP though the addition of a vector $Y \in \mathbb{R}^{H,W,C}$ where each $Y_{i,j,k}$ is an independent random variable following a Laplace distribution with a scaling parameter 
\end{lemma}

\begin{equation}
\sigma_{i,j,k}  = \frac{r^{i,j,k}_{max} - r^{i,j,k}_{min}}{\epsilon_{i,j,k}},
~~~~~~{\rm where} \sum\limits_{i,j,k}\epsilon_{i,j,k} = \epsilon
\label{dpdef} 
\end{equation}

With the setting in Lemma \ref{lemma}, Equation \ref{newdp} can be guaranteed. The proof has been attached in Appendix.

\subsubsection{Differential Privacy Perturbation Module.}
\label{DPmod}

The size of the output of the frequency domain transformation module is $[H, W, C]$. 
We need a noise matrix of the same size to mask the frequency features. 
Due to the properties of DCT, most energies are gathered in small areas. 
Thus, the frequency features have very different importance for face recognition. 
In order to protect face privacy with the best possible face recognition accuracy, we use a learnable privacy budgets allocation method in the module. 
It allows us to assign more privacy budgets to locations that are important for face recognition. 

To achieve the idea of learnable privacy budgets, we utilize the setting in Lemma 1. 
We first initialize learnable privacy budget assignment parameters with the same size as the frequency features and put them into the softmax layer. 
Then, we multiply each element of the output of the softmax layer by $\epsilon$.
Because the sum of the output after the softmax layer is 1, no matter how the learnable budget assignment parameters are changed, the total privacy budget always equals $\epsilon$. 
According to Lemma 1, if we add a vector $Y \in \mathbb{R}^{H,W,C}$ to the frequency features where each $Y_{i,j,k}$ is an independent random variable following a Laplace distribution with a scaling parameter $\sigma_{i,j,k}  = \frac{r^{i,j,k}_{max} - r^{i,j,k}_{min}}{\epsilon_{i,j,k}}$, then we can ensure that this feature is protected by $\epsilon-DP$. 
To get the sensitivities of the frequency features of facial images, we transferred all the images in VGGFace$2$~\cite{cao2018vggface2} and refined MS1MV$2$ into the frequency domain. 
Then we obtained the maximum values and minimum values at each position. 
Sensitivities will equal the value of MAX - MIN. 
By now, we have prepared all the scaling parameters of Laplace noise. 
We sample the Laplace distribution according to the parameters and add them to the frequency features. 
The masked frequency features will be the output of the differential privacy perturbation module and transmitted to the face recognition model. 
The learnable budget assignment parameters are learned based on the loss function of the face recognition model and get the best allocation scheme that guarantees recognition accuracy. For face recognition module, we use ArcFace \cite{deng2019arcface} as loss function and ResNet50 \cite{he2016deep} as backbone.

\section{Experiment}
\subsection{Datasets}
\noindent
We use VGGFace$2$~\cite{cao2018vggface2} that contains about 3.31M images of 9131 subjects for training.
We extensively test our method on several popular face recognition benchmarks, including five small testing datasets and two general large-scale benchmarks. 
LFW~\cite{huang2008labeled} contains $13233$ web-collected images from $5749$ different identities.
CFP-FP~\cite{sengupta2016frontal}, CPLFW~\cite{zheng2017cross}, CALFW~\cite{zheng2017cross} and AgeDB~\cite{moschoglou2017agedb} utilize the similar evaluation metric of LFW to test face recognition with various challenges, such as cross pose, cross age.IJB-B~\cite{whitelam2017iarpa} and IJB-C~\cite{maze2018iarpa} are two general large-scale benchmarks.
The IJB-B dataset contains $12,115$ templates with $10,270$ genuine matches and $8$M impostor matches.
The IJB-C dataset is a further extension of IJB-B, having $23,124$ templates with $19,557$ genuine matches and $15,639$K impostor matches.
\subsection{Implementation Details}
\noindent
Each input face is resized to $112 \times 112$. 
After the frequency domain transformation, the input channel $C$ is set to 189.
We set the same random seed in all experiments.
Unless otherwise stated, we use the following setting.
We train the baseline model on ResNet50~\cite{he2016deep} backbone.
For our proposed model, we first convert the raw input RGB image to BDCT coefficients using some functions in TorchJPEG~\cite{ehrlich2020quantization}.  
Secondly, we process the BDCT coefficients using the proposed method.
 We calculate the sensitivity among the whole training dataset.
The initial values of the learnable budget allocation parameters are set to be $0$ so that the privacy budget of each pixel is equal in the initial stage.
The whole model is trained from scratch using the SGD algorithm for $24$ epochs, and the batch size is set to be $512$.
The learning rate of the learnable budget allocation parameters and the backbone parameters is 0.1.
The momentum and the weight-decay are set to be $0.9$ and $5\mathrm{e}{-4}$, respectively.
We conducted all the experiments on $8$ NVIDIA Tesla V$100$ GPU with the PyTorch framework.
We divide the learning rate by $10$ at $10$, $18$, $22$ epochs.
For ArcFace~\cite{deng2019arcface}, we set $s=64$ and $m=0.4$.
For CosFace~\cite{wang2018cosface}, we set $s=64$ and $m=0.35$.

\subsection{Comparisons with SOTA Methods}
\label{subsection:Compare}
\subsubsection{Settings for Other Methods.}
To evaluate the effectiveness of the model, we compare it with five baselines:
\textbf{(1) ArcFace~\cite{deng2019arcface}}: The model is ResNet50 equipped with ArcFace, which is the simplest baseline with the original RGB image as input and introduces an additive angular margin inside the target cosine similarity.
\textbf{(2) CosFace~\cite{wang2018cosface}}: The model is ResNet50 equipped with CosFace, which is another baseline with the original RGB image as input and subtracts a positive value from the target cosine similarity.
\textbf{(3) PEEP~\cite{chamikara2020privacy}}: This method is the first to use DP in privacy-preserving face recognition. 
We reproduce it and run it on our benchmarks. 
Due to a large amount of training data, half of the data is selected for each ID to calculate the eigenface. 
The privacy budget $\epsilon$ is set to 5.
\textbf{(4) Cloak~\cite{mireshghallah2021not}}: We run its official code. 
Note that when training large datasets, the privacy-accuracy parameter is adjusted according to our experiment setting, which is set to 100. 
\textbf{(5) InstaHide~\cite{huang2020instahide}}: This method incorporates mix-up to solve the privacy-preserving problem. 
We adapt it to face privacy-preserving problems. 
We set $k$ to 2 and adopt an inside-dataset scheme that mixes each training image with random images within the same private training dataset.

\begin{table*}
\begin{center}
\resizebox{0.9\textwidth}{!}{
\begin{tabular}{lllllllll}
    \toprule
    Method (\%) & Privacy-Preserving & LFW & CFP-FP& AgeDB-30 & CALFW & CPLFW & IJB-B(1e-4) & IJB-C(1e-4) \\
    \midrule
    ArcFace (Baseline) & No  & 99.60 & 98.32 & 95.88	& 94.16	& 92.68 & 91.02 & 93.25 \\
    CosFace (Baseline) & No & 99.63 & 98.52 & 95.83 & 93.96 & 93.30 & 90.79 & 93.14 \\
    PEEP & Yes & 98.41 & 74.47 & 87.47 & 90.06 & 79.58 & 5.82 &  6.02\\
    Cloak & Yes & 98.91 & 87.97 & 92.60 & 92.18 & 83.43 & 33.58  & 33.82\\
    InstaHide & Yes & 96.53 & 83.20 & 79.58 & 86.24 & 81.03 & 61.88 & 69.02 \\
    \textbf{Ours}, Arcface ($\epsilon_{mean}=0.5$) & Yes & \textbf{99.48} & \textbf{97.20} & \textbf{94.37} & \textbf{93.47} & 90.6 & 89.33 & \textbf{91.22} \\
    \textbf{Ours}, Cosface ($\epsilon_{mean}=0.5$) & Yes & 99.47 & 97.16 & 94.13 & 93.36 & \textbf{90.88} & \textbf{89.37} & 91.21 \\
    \bottomrule
\end{tabular}
}

\caption{Comparison of the face recognition accuracy among different privacy-preserving face recognition methods.}
\label{tableacc}
\end{center}
\end{table*}

\subsubsection{Results on LFW, CFP-FP, CPLFW, CALFW, AgeDB.}
The results of the comparison with other SOTA methods can be seen in Tab.\ref{tableacc}. For our method, the accuracy on all five datasets is close. 
Our method has a similar performance to the baselines on LFW and CALFW, with only an average drop of 0.14$\%$ and 0.65$\%$. 
For CFP-FP, AgeDB, and CPLFW, our method has an average drop of 2.5$\%$ compared with the baseline. 
We believe this is because the images in these datasets have more complex poses and are therefore inherently less robust and more susceptible to interference from noise. 
However, our method still has a considerable lead in performance on these datasets compared to other SOTA privacy-preserving methods. 
In particular, on the CFP-FP dataset, other privacy-preserving methods have accuracy losses of more than 10$\%$, but we still perform well.
\subsubsection{Results on IJB-B and IJB-C.}
We also compare our method with baseline and other SOTA privacy-preserving methods over IJB-B and IJB-C. 
As shown in Tab.\ref{tableacc} and Fig.\ref{figfirst}, our method has a very similar performance compared to the baseline. 
Under different false positive rates, the true positive rate of our method is still at a high level. 
However, the other SOTA methods do not perform well in this area. 
Their true-positive rate is much lower than the baseline with the lower false-positive rate.

\subsection{Privacy Attack}
\subsubsection{White-box Attacking Experiments.}
\label{parawhite}

\begin{figure}
\begin{minipage}[t]{0.5\linewidth}
\centering
\includegraphics[width=0.9\columnwidth]{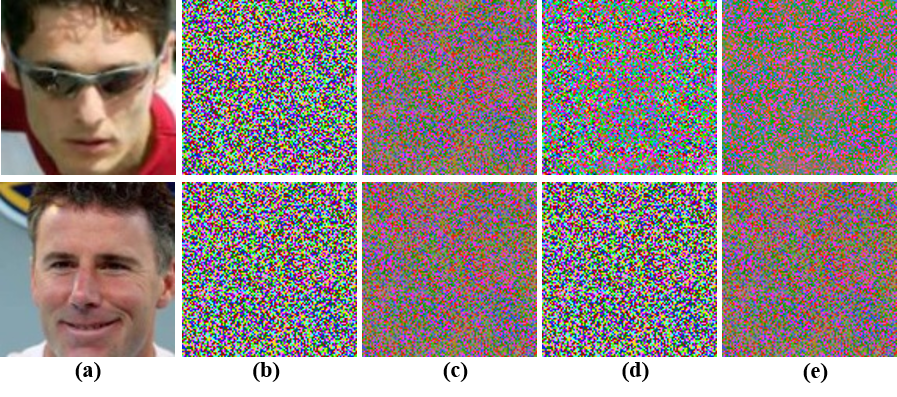}
\begin{minipage}[t]{0.98\linewidth}
   \caption{White-box attack for our method. (a) Raw image. (b) IDCT with $\epsilon$=0.5. (c)Denoising image of (b). (d) IDCT with $\epsilon$=100. (e) Denoising image of (d).
   }
\label{figwhitenodc}
\end{minipage}
\end{minipage}
\begin{minipage}[t]{0.5\linewidth}
    \centering
    \includegraphics[width=0.9\columnwidth]{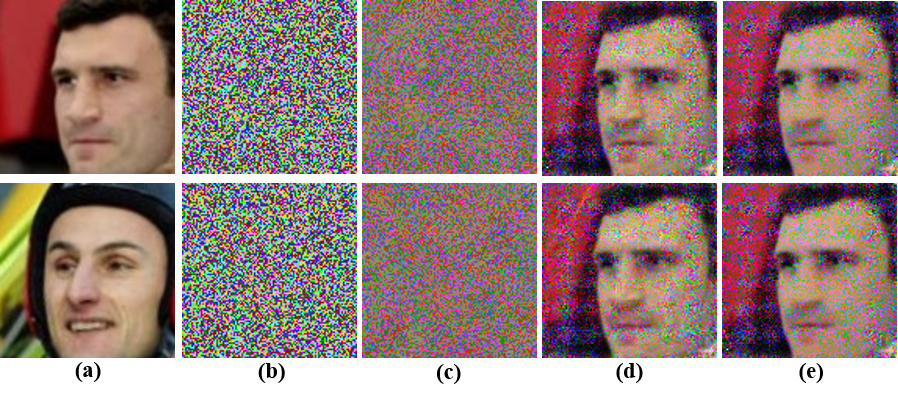}
    \begin{minipage}[t]{0.98\linewidth}
    \caption{White-box attack for our method with guessed DC. For both line, we add the DC of the upper user. (a) Raw image. (b) IDCT with $\epsilon$=0.5. (c) Denoising image of (b). (d) IDCT with $\epsilon$=100. (e) Denoising image of (d).
   }
   
\label{figwhitedc}
\end{minipage}
\end{minipage}
\end{figure}

We assume that the attacker already knows all our operations in the white-box attack section. 
Therefore he will perform an inverse DCT (IDCT) operation on the transmitted data. Since our operation to remove DCs is practically irreversible, we assume he fills all DCs to 0. 
We set the privacy budget to 0.5 and 100, respectively, to better demonstrate the effect. However, in practice, we do not recommend setting the privacy budget as large as 100. 
Further, we denoise the images after IDCT. 
Here we use non-local means denoising~\cite{buades2005non} as the denoising method. 
As shown in Fig.\ref{figwhitenodc}, after losing the information of DC, it is difficult to show the original facial information in the recovered figure of the white box attack. 
Even if the privacy budget is as large as 100, we cannot obtain valid information about the user's facial features from the recovered image. 
Moreover, at this time, the denoising method also cannot work effectively because the IDCT image is full of noise.

To further demonstrate our approach's inability to provide valid information about the user's facial features, we assume that the attacker knows this data originates from a specific dataset. 
He makes certain guesses about the user to whom the data belongs and adds the DC of the guessed user to the data. 
The corresponding results are shown in Fig.\ref{figwhitedc}. 
We assumed that the attacker guessed exactly the corresponding user in the above row. However, with the privacy budget set to 0.5, the recovery image still does not reveal any facial information. 
In the following row, we assume that the data belong to another user. However, the attacker still guesses that the data belongs to the same user as in the above row. 
In other words, the attacker guesses the wrong data source and adds the wrong DC to it. 
The recovery image does not give any information to the attacker about whether he guesses the right user to whom the data belongs. 
In this way, we can say that our method can protect the user's privacy well.

\subsubsection{Black-box Attacking Experiments.}
\label{blackattack}
\begin{figure}
  \begin{minipage}[t!]{0.45\linewidth}
  \centering
\includegraphics[width=1\columnwidth]{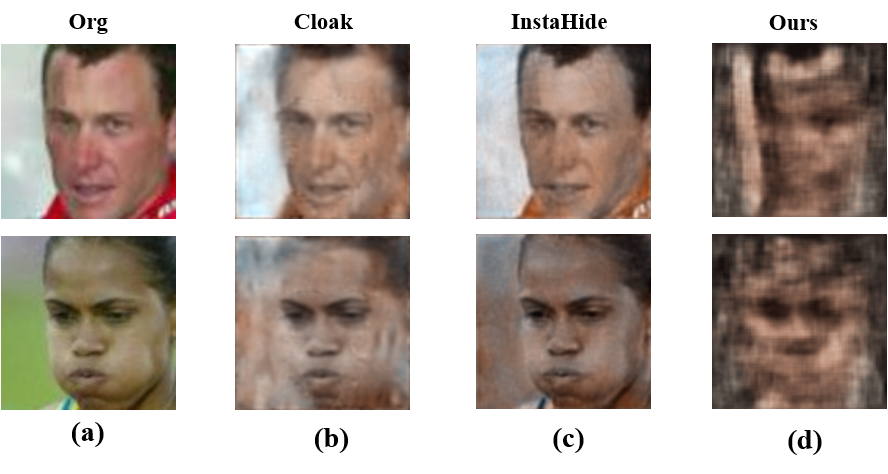}
   \caption{Visualization of different privacy-preserving methods under black-box attack.
   }
\label{figblack}
\end{minipage}
\begin{minipage}[t!]{0.55\linewidth}
\centering
\resizebox{1\columnwidth}{!}{
\begin{tabular}{l|c|ccccc}
\toprule
Metric  &Method   & LFW  & CFP-FP  & CPLFW  & AgeDB   & CALFW \\ \midrule
           &Cloak  & 20.292 & 20.069 & 19.705 & 19.946 & 20.327 \\
PSNR(db)   &InstaHide  & 23.519 & 22.807 & 22.546 & 23.271 & 23.010 \\
           &\textbf{Ours}  &\textbf{14.281}  &\textbf{13.145}  &\textbf{13.515}  &\textbf{13.330}  &\textbf{13.330}\\
           \midrule
           &Cloak  & 0.564 & 0.464 & 0.578 & 0.574 & 0.526 \\
Similarity &InstaHide  & 0.729 & 0.649 & 0.732 & 0.737 & 0.693 \\
           &\textbf{Ours}  &\textbf{0.214}  &\textbf{0.175} &\textbf{0.264}  &\textbf{0.250}  &\textbf{0.202}   \\
           \bottomrule
\end{tabular}
}
\begin{minipage}[t]{0.97\linewidth}
\caption{PSNRs and similarities between the original images and the ones recovered from output of different privacy-preserving methods. The lower the value is, the better the privacy-preserving method is.}
\label{tableblackfs}
\end{minipage}
\end{minipage}
\end{figure}
In this section, we analyze the privacy-preserving reliability of the proposed method from the perspective of a black-box attack.
A black box attack means that the attacker does not know the internal structure and parameters of the proposed model.
However, attackers can collect large-scale face images from websites or other public face datasets.
They can obtain the processed inputs by feeding those data to the model.
Subsequently, they can train a decoder to map the processed inputs to the original face images.
Finally, attackers can employ the trained decoder to recover the user's face image.
Under these circumstances, we use UNet~\cite{ronneberger2015u} as our decoder to reconstruct original images from processed images.
In the training phase, we use an SGD optimizer with a learning rate of 0.1 with 10 epochs, and the batch size is set to 512.
As shown in Fig.\ref{figblack}, we compare our method with other methods on reconstructed images. 
For Cloak, the face is still evident since the added noise only affects the face's background.
For InstaHide, most encrypted images can be recovered under our setting.
For our method, the reconstructed images have been blurred.
The facial structure of recovered images has been disrupted with an average privacy budget of 0.5. 

In addition, Fig.\ref{tableblackfs} shows some quantitative results to illustrate the effectiveness of our method further.
The reported results correspond to the results in Tab.\ref{tableacc}.
We first compare the PSNR between the original image and the one reconstructed. PSNR is often used to measure the reconstruction quality of lossy compression. It is also a good measure of image similarity. 
As we can find in the first row of Fig.\ref{tableblackfs}, compared with other methods, our method has lower PSNR and higher recognition accuracy.
Furthermore, we also evaluate the average Feature Similarity across five small data sets in order to show the privacy-preserving ability of the model at the feature level, as shown in the second row of Fig.\ref{tableblackfs}.
Precisely, we feed recovered RGB images to a pre-trained Arcface backbone.
Then we can get new feature embedding to perform feature similarity calculation with origin embedding.
The smaller the similarity, the more recovery-resistant the privacy-preserving method is at the feature level.
As we can see, when the average privacy budget is chosen to be lower than 1, the similarities are much smaller than the other methods.

\begin{figure}[t!]
\centering
\subfigure{
\begin{minipage}[t]{0.5\linewidth}
\centering
\includegraphics[width=0.7\linewidth]{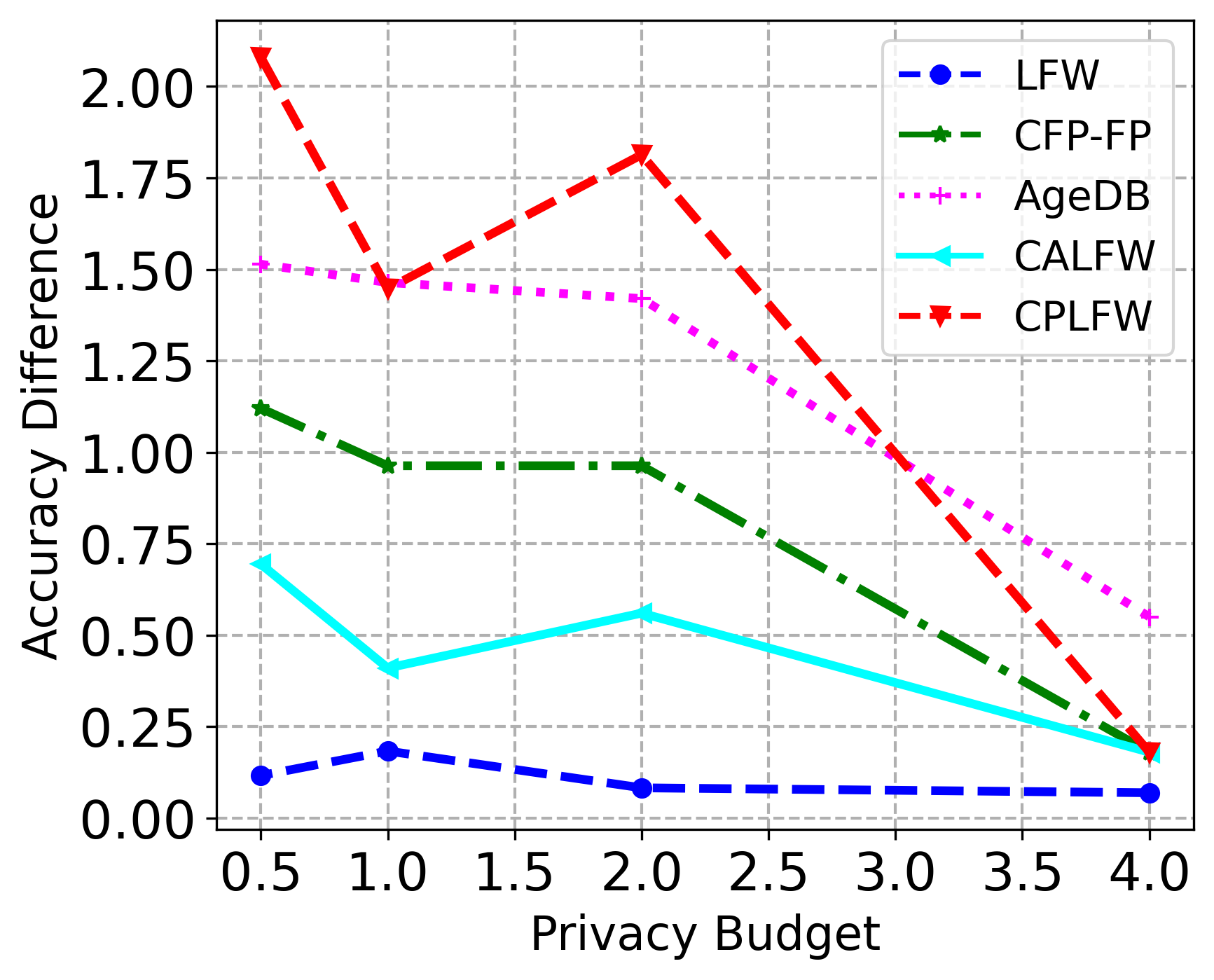}
\end{minipage}%
}%
\subfigure{
\begin{minipage}[t]{0.47\linewidth}
\centering
\includegraphics[width=0.7\linewidth]{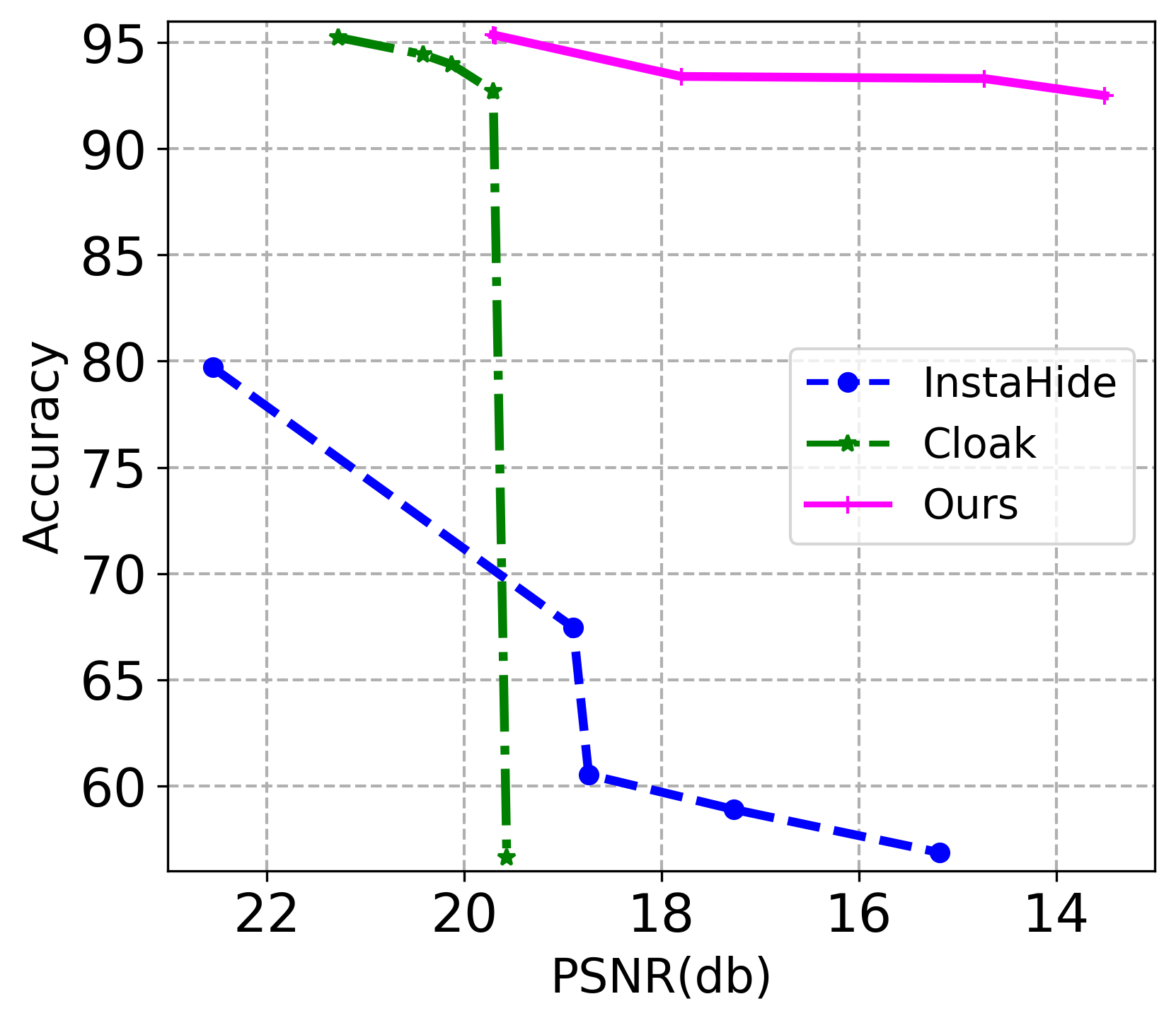}
\end{minipage}%
}
\centering
\caption{ \textbf{Left}: The recognition accuracy among different settings of privacy budget under different data set. The lower the value is, the less impact the method has on recognition accuracy. \textbf{Right}: Trade-off abilities of different approaches. The position in the upper right-hand corner indicates greater resilience to recovery while maintaining accuracy.}
\label{figall}
\end{figure}

We further experimented with the trade-off ability of different approaches for privacy and accuracy, as shown in Fig.\ref{figall}.
Cloak does not protect the privacy of the face to a high degree, so the PSNR of the recovered image under black-box attack and the original image can hardly be lower than 19.
In contrast, InstaHide can protect face privacy quite well as the number of mixed images increases. However, the resulting loss of accuracy is very heavy.
For our approach, depending on the choice of the average privacy budget, the degree of face privacy protection can be freely chosen, and the loss of accuracy can be controlled to a small extent.
In summary, our method is more resistant to recovery than other methods, which means a more robust privacy-preserving ability.

\subsection{Ability to protect the privacy of training data.}
Unlike other privacy-preserving methods, we can protect privacy during the inference and training stages. 
After the first training, we can get fixed privacy budget allocation parameters. 
We perform a DCT operation on the existing raw face recognition dataset and add noise according to the privacy budget allocation parameters. 
Thus, the original privacy leaked dataset is transformed into a privacy-protected dataset. Moreover, we experimentally demonstrate that the transformed dataset can still be used to train face recognition tasks with high accuracy. 
As shown in Tab.\ref{tabledataset}, the model trained using privacy-preserving datasets still has high accuracy.
Notably, even though we trained fixed privacy assignment parameters using VGGFace2 and transformed MS1M using it, the transformed privacy-preserving MS$1$M retained high usability. It demonstrated the versatility and separability of the privacy budget allocation parameters.
The corresponding results are shown in the fourth row.

\begin{table}[t!]
\centering
\resizebox{0.8\columnwidth}{!}{
\begin{tabular}{lllllll}
    \toprule
    Method (\%) & LFW & CFP-FP& AgeDB-30 & CALFW & CPLFW  \\
    \midrule
    VGGFace2  & 99.60 & 98.32 & 95.88 & 94.16 & 92.68  \\
    MS$1$M  & 99.76 & 97.94 & 98.00 & 96.10 & 92.30 \\
    Privacy-Preserved VGGFace2  & 99.68 & 97.88 & 95.85 & 93.97 & 92.11 \\
    Privacy-Preserved MS$1$M  & 99.73 & 96.94 & 97.96 & 95.96 & 91.73 \\
    \bottomrule
\end{tabular}
}
\caption{Comparison of recognition accuracy training with different datasets. \textbf{VGGFace2/MS$1$M}: Training with the original VGGFace2/MS$1$M. \textbf{Privacy-Preserved VGGFace2/MS$1$M}: Training with the dataset that transformed from the original VGGFace2/MS$1$M. They used the same fixed privacy budget allocation parameters learned from VGGFace2 and $\epsilon=2$.}
\label{tabledataset}
\end{table}

\subsection{Ablation Study}
\label{ablation}

\subsubsection{Effects of transformation to the frequency domain.}
To demonstrate the effects of transformation to the frequency domain, we directly input the raw RGB facial image to the differential privacy perturbation module. 
Here we choose $\epsilon_{mean} = 0.5$. 
As shown in Tab.\ref{tableaccabl}, the model without transformation to frequency domain has a much lower accuracy even with the same setting for other parts.

\begin{figure}[htb]
\begin{center}
\includegraphics[width=0.7\columnwidth]{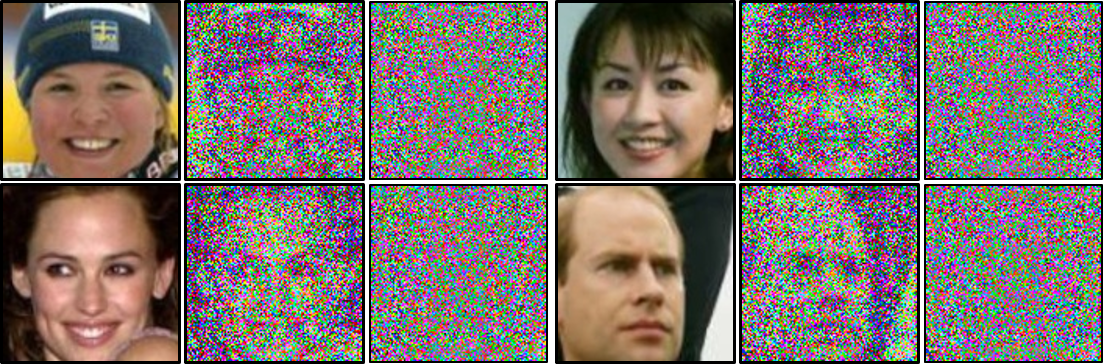}
\end{center}
   \caption{From left to right, the original image, the IDCT image without DC removal and the IDCT image with dc removal.
   }
\label{figwithDCs}
\end{figure}

\subsubsection{Effects of removing DCs.}
To show the effects of removing DCs, we train models with and without removing DCs for comparison. 
We chose a larger privacy budget to make the difference between removing DC and not removing DC more visible. 
Here we set the budget equal to 20. 
By comparing the IDCT visualization of masked images without removing DCs shown in Fig.\ref{figwithDCs}, we can see that the removal of DCs has a good effect on facial information protection.

\begin{table}[t]
\centering
\resizebox{0.8\columnwidth}{!}{
\begin{tabular}{lllllll}
    \toprule
    Method (\%) & LFW & CFP-FP& AgeDB-30 & CALFW & CPLFW  \\
    \midrule
    ArcFace (Baseline)  & 99.60 & 98.32	& 95.88	& 94.16	& 92.68  \\
    BDCT-DC  & 99.68 & 98.3 & 95.93 & 94.25 & 93.28 \\
    BDCT-NoDC   & 99.48 &	98.39	& 95.73 & 94.35 & 92.92 \\
    RGB ($\epsilon_{mean}=0.5$)  & 84.78 & 63.87 & 73.66 & 76.85 & 63.08 \\
    \textbf{Ours} ($\epsilon_{mean}=0.5$)  & \textbf{99.48} & \textbf{97.20} & \textbf{94.37} & \textbf{93.47} & \textbf{90.6} \\
    \bottomrule
\end{tabular}
}
\caption{Comparison of the face recognition accuracy among methods with or without privacy protection. \textbf{BDCT-DC/BDCT-NoDC}: The baseline model trained on BDCT coefficients with/without DC. \textbf{RGB}: The model trained on the original image with learnable privacy budgets.}
\label{tableaccabl}
\end{table}

\subsubsection{Effects of using learnable DP budgets.}
To demonstrate the effects of using learnable DP budgets, we compare the accuracy of using learnable DP budgets and using the same DP budget for all elements. 
We set the privacy budget to be 0.5 at every element and test its accuracy using a pre-trained model with noiseless frequency domain features as input. 
The accuracy over LFW is only 65.966$\%$, which is much lower than the one with learnable DP budgets, shown in Tab.\ref{tableaccabl}.

\subsubsection{Effects of choosing different DP budgets.}
To show the effect of choosing different privacy budgets, we chose different privacy budgets and tested their loss of accuracy relative to the baseline. 
The results of the tests are presented in Fig.\ref{figall}.
As we have seen, the smaller the privacy budget is set, the higher the accuracy loss will be. 
This result is in line with the theory of differential privacy that a smaller privacy budget means more privacy-preserving and has a correspondingly higher accuracy loss.
In our approach, the smaller privacy budget also provides stronger privacy protection, which is proven in Section  \ref{blackattack}.

\section{Conclusions}
In this paper, we propose a privacy-preserving approach for face recognition. It provides a privacy-accuracy trade-off capability. Our approach preserves image privacy by transforming the image to the frequency domain and adding random perturbations. It utilizes the concept of differential privacy to provide strong protection of face privacy. It regulates the ability of privacy protection by changing the average privacy budget. It is lightweight and easily compatible to be easily added to existing face recognition models.
Moreover, it is shown experimentally that our method can fully protect face privacy under white-box attacks and maintain similar accuracy as the baseline. The masked images can defend against the black-box recovery attack of UNet. These show that our method performs far better than other SOTA face recognition protection methods.

\bibliographystyle{splncs04}
\bibliography{eccv22}
\end{document}